\documentclass[10pt,twocolumn,letterpaper]{article}
\pdfoutput=1

\usepackage{iccv}
\usepackage{times}
\usepackage{epsfig}
\usepackage{graphicx}
\usepackage{amsmath}
\usepackage{comment}
\usepackage{amssymb}
\usepackage[vlined,boxed,commentsnumbered,ruled,linesnumbered]{algorithm2e}
\usepackage{multirow}

\usepackage{booktabs,array,xcolor,colortbl,epsfig,graphicx,subfigure}
\usepackage{stfloats}

\usepackage[breaklinks=true,bookmarks=false]{hyperref}

\iccvfinalcopy % *** Uncomment this line for the final submission

\def\iccvPaperID{41} % *** Enter the ICCV Paper ID here

\ificcvfinal\fi
\begin{document}

%%%%%%%%% TITLE
%\title{Learning Permutation Loss based Cross-Graph Embedding Networks
%\\for Deep Graph Matching}
\title{Learning Combinatorial Embedding Networks for Deep Graph Matching}

\author{Runzhong~Wang\textsuperscript{1,2} \qquad Junchi~Yan\textsuperscript{1,2 }\thanks{Corresponding author. This work is supported by National Key Research and Development Program of China (2016YFB1001003), STCSM (18DZ1112300), NSFC (61602176). } \qquad Xiaokang~Yang\textsuperscript{2} \\
\textsuperscript{1} Department of Computer Science and Engineering, Shanghai Jiao Tong University \\
\textsuperscript{2} MoE Key Lab of Artificial Intelligence, AI Institute, Shanghai Jiao Tong University\\
{\tt\small \{runzhong.wang,yanjunchi,xkyang\}@sjtu.edu.cn}
%\and
}

\maketitle
\thispagestyle{empty}

%%%%%%%%% ABSTRACT
%It has been a long-standing problem due to both its usefulness in various applications and its NP-complete nature.
\begin{abstract}
Graph matching refers to finding node correspondence between graphs, such that the corresponding node and edge's affinity can be maximized. In addition with its NP-completeness nature, another important challenge is effective modeling of the node-wise and structure-wise affinity across graphs and the resulting objective, to guide the matching procedure effectively finding the true matching against noises. To this end, this paper devises an end-to-end differentiable deep network pipeline to learn the affinity for graph matching. It involves a supervised permutation loss regarding with node correspondence to capture the combinatorial nature for graph matching. Meanwhile deep graph embedding models are adopted to parameterize both intra-graph and cross-graph affinity functions, instead of the traditional shallow and simple parametric forms e.g. a Gaussian kernel. The embedding can also effectively capture the higher-order structure beyond second-order edges. The permutation loss model is agnostic to the number of nodes, and the embedding model is shared among nodes such that the network allows for varying numbers of nodes in graphs for training and inference. Moreover, our network is class-agnostic with some generalization capability across different categories. All these features are welcomed for real-world applications. Experiments show its superiority against state-of-the-art graph matching learning methods.
  
  %Our network has its real-world applicability in that it can be trained by graph pairs for each containing different numbers of nodes.
\end{abstract}
% In this paper, we propose the first combination supervised end-to-end deep graph matching model adopting graph embedding layers, especially cross-graph embedding.
\section{Introduction and Preliminaries}
\label{sec:intro}
Graph matching (GM) refers to establishing node correspondences between two or among multiple graphs. Graph matching incorporates both the unary similarity between nodes and pairwise~\cite{ChoECCV10, GoldPAMI96} (or even higher-order~\cite{LeeCVPR11,NgocCVPR15,YanCVPR15}) similarity between edges from separate graphs to find a matching such that the similarity between the matched graphs is maximized. By encoding the high-order geometrical information in the matching procedure, graph matching in general can be more robust to deformation noise and outliers. For its expressiveness and robustness, graph matching has lied at the heart of many computer vision applications e.g. visual tracking, action recognition, robotics, weak-perspective 3-D reconstruction -- refer to \cite{VentoCV12} for a more comprehensive survey on graph matching applications.
% This is in contrast to the point based methods e.g. RANSAC~\cite{FischlerCACM81} and iterative closet point (ICP)~\cite{ZhangIJCV94} whereby such edge-to-edge information is not explicitly modeled.

Due to its high-order combinatorial nature, graph matching is in general NP-complete~\cite{Gare90NPComplete} such that researchers employ approximate techniques to seek inexact solutions. For the classic setting of two-graph matching between graphs $\mathcal{G}_1$, $\mathcal{G}_2$, the problem can be written by the following general quadratic assignment programming (QAP) problem~\cite{LoiolaEJOR07}:
\begin{align}\label{eq:lawler_qap}
&J(\mathbf{X}) = \text{vec}(\mathbf{X})^\top\mathbf{K}\text{vec}(\mathbf{X}),\\\notag
&\mathbf{X}\in \{0,1\}^{N\times N}, \quad\mathbf{X}\mathbf{1}= \mathbf{1}, \quad \mathbf{X}^\top\mathbf{1}\leq\mathbf{1}
\end{align}
%$$\min_{\mathbf{X}\in \{0,1\}^{n_1\times n_2}}\text{vec}(\mathbf{X}^\top)\mathbf{K}\text{vec}(\mathbf{X})$$,
where $\mathbf{X}$ is a permutation matrix indicating the node correspondence, and $\mathbf{K}\in \mathbb{R}^{N^2\times N^2}$ is the so-called affinity matrix~\cite{LeordeanuICCV05} whose diagonal elements and off-diagonal ones encode the node-to-node and edge-to-edge affinity between two graphs, respectively. One popular embodiment of $\mathbf{K}$ in literature is $\mathbf{K}_{ia,jb}=\exp\left(\frac{(\mathbf{f}_{ij}-\mathbf{f}_{ab})^2}{\sigma^2}\right)$ where $\mathbf{f}_{ij}$ is the feature vector of the edge $ij$, which can also incorporate the node similarity when node index $ia=jb$.  

Eq.~(\ref{eq:lawler_qap}) is called Lawler's QAP~\cite{LawlerMS63}. It can incorporate other forms e.g. Koopmans-Beckmann's QAP~\cite{LoiolaEJOR07}:
\begin{equation}
J(\mathbf{X})=\text{tr}(\mathbf{X}^\top\mathbf{F}_1\mathbf{X}\mathbf{F}_2)+\text{tr}(\mathbf{K}_p^\top\mathbf{X})
\end{equation}
where $\mathbf{F}_1\in \mathbb{R}^{N\times N}$, $\mathbf{F}_2\in \mathbb{R}^{N\times N}$ are weighted adjacency matrices of graph $\mathcal{G}_1$, $\mathcal{G}_2$ respectively, and $\mathbf{K}_p$ is the node-to-node affinity matrix. Its connection to the Lawler's QAP can be established by setting $\mathbf{K}=\mathbf{F}_2\otimes\mathbf{F}_1$.
\begin{comment}
Another popular formulation is the so-called factorized graph matching model \cite{ZhouPAMI16}, which shows how to factorize affinity matrix $\mathbf{K}$ as a Kronecker product of smaller matrices. For concise, here we write the undirected version~\cite{ZhouPAMI16}:
\begin{align}
&\mathbf{K}=(\mathbf{H}_2\otimes\mathbf{H}_1)\text{diag}(\text{vec}(\mathbf{L}))(\mathbf{H}_j\otimes\mathbf{H}_i)^\top\\\notag
\text{where}\quad&\mathbf{H}_k=[\mathbf{G}_k,\mathbf{I}_{n_k}]\in\{0,1\}^{n_k\times(m_k+n_k)},\quad k=i,j\\\notag
&\mathbf{L}= \left[\begin{array}{ll} \mathbf{K}^{q} & -\mathbf{K}^{q}\mathbf{G}_{j}^\top\\
-\mathbf{G}_{i}\mathbf{K}^{q}  & \mathbf{G}_{i}\mathbf{K}^{q}\mathbf{G}_{j}^\top+\mathbf{K}^{p}\end{array}\right]
\end{align}
where $n_i$ and $m_i$ is the number of nodes and edges in graph $\mathcal{G}_i$ respectively and $\otimes$ is the Kronecker product operation between matrices. $\mathbf{K}^{p}\in\mathbb{R}^{n_i\times n_j}$ denotes the node affinity matrix, and $\mathbf{K}^q\in\mathbb{R}^{m_i\times m_j}$ for the edge affinity matrix. The graph structure is specified by the node-edge incidence matrix $\mathbf{G}\in\mathbb{R}^{n\times m}$ such that the non-zero elements in each column of $\mathbf{G}$ indicate the starting and ending nodes in the corresponding edge. The factorization provides a taxonomy for GM and reveals the connection among several methods. Readers are referred to \cite{ZhouPAMI16} for greater details.
\end{comment}

Beyond the second-order affinity modeling, recent methods also explore the way of utilizing higher-order affinity information. Based on tensor marginalization as adopted by several hypergraph matching works \cite{ChertokPAMI10,DuchennePAMI11,YanCVPR15,ZassCVPR08}:
\begin{gather}\label{eq:formulationB}
\mathbf{x}^{*}=\arg \max (\mathbf{H}\otimes_1 \mathbf{x}\otimes_2 \mathbf{x}\ldots\otimes_m \mathbf{x})\quad s.t. \\\notag
\mathbf{X}\mathbf{1}= \mathbf{1}, \mathbf{X}^\top\mathbf{1}\leq\mathbf{1}, \mathbf{x} = \text{vec}(\mathbf{X}) \in \{0,1\}^{N^2\times 1}
\end{gather}
where $m$ is the affinity order and $\mathbf{H}$ is the $m$-order affinity tensor whose element encodes the affinity between two hyperedges from the graphs. $\otimes_k$ is the tensor product \cite{LeeCVPR11}. Readers are referred to Sec. 3.1 in \cite{DuchennePAMI11} for details on tensor multiplication. The above works all assume the affinity tensor is invariant w.r.t. the index of the hyperedge pairs.

%In addition with the above affinity formulations for two-graph matching, there are also a considerable amount of recent efforts on multiple graph matching approaches. In fact, 

%There are also multi-graph matching approaches mostly either transform the problem into a two-graph matching QAP problem in each iteration \cite{YanPAMI16,yan2015consistency}, or first solve a two-graph matching problem to obtain the putative matchings for further smoothing in post-processing step \cite{chen2014near,HuangSGP13,PachauriNIPS13,wang2018multi}.

\begin{figure*}[t!]
    \centering
    \includegraphics[width=\textwidth]{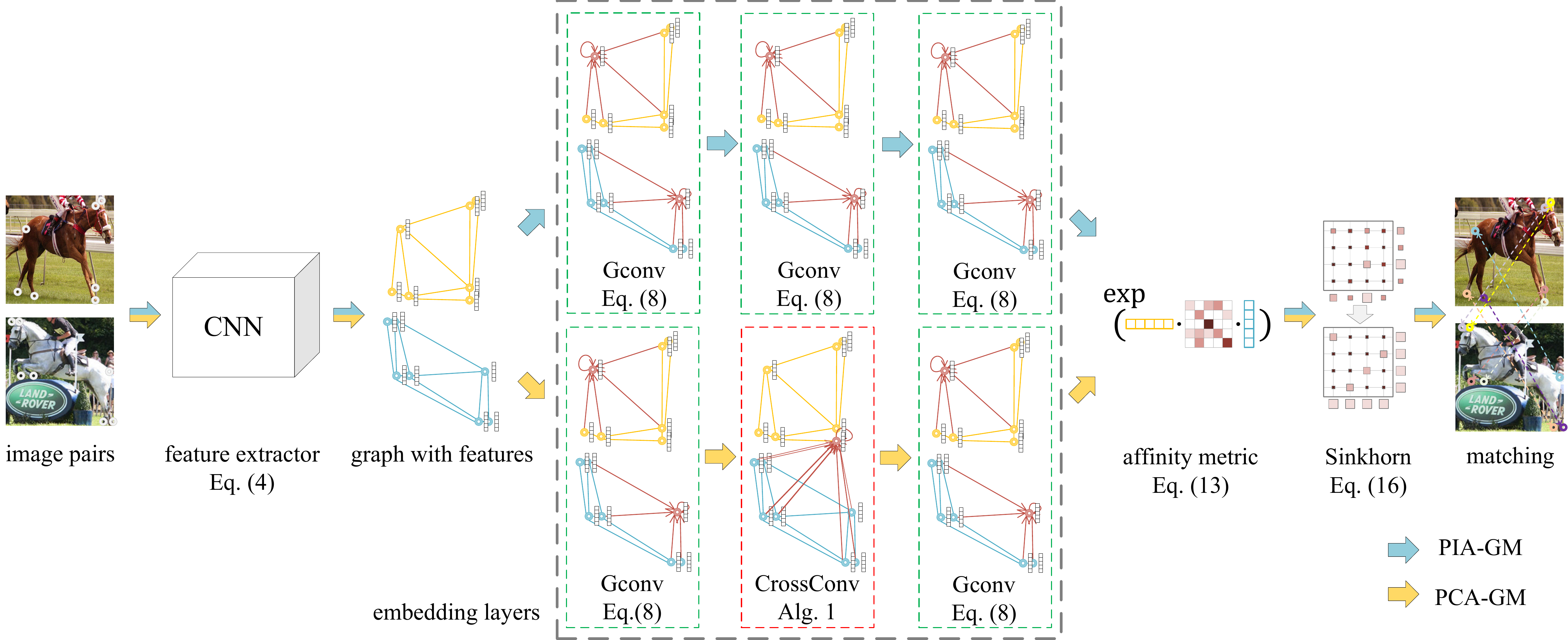}
    \vspace{-10pt}
    \caption{Overview of our proposed permutation based intra-graph affinity (PIA-GM) and cross-graph affinity (PCA-GM) approaches for deep combinatorial learning of graph matching. The CNN features are extracted from image pairs followed by node embedding and Sinkhorn operation for matching. The CNN model, embedding model and affinity metric are all learnable in an end-to-end fashion.}
    \vspace{-10pt}
    \label{fig:overview}
\end{figure*}

All the above studies show the generality and importance of the affinity model for graph matching. However, traditional affinity methods mostly rely on a predefined affinity function (or distance), e.g. a Gaussian kernel with Euclid distance in the node and edge feature space. We believe that such a predefined parametric affinity model has limited flexibility to capture the structure of a real-world matching task, whereby the affinity metric can be arbitrary and call for models with enough high capacity to approximate. This challenge is more pronounced in the presence of noise and outliers which are ubiquitous in practical settings. Based on an inappropriate affinity model, the matching solver can be more struggling as the global optimum regarding with the affinity model may even not correspond to the ground truth matching solution -- due to the biased affinity objective function as input for combinatorial optimization.

Hence it calls for effective affinity modeling across graphs. It is orthogonal to the major line of previous efforts on devising combinatorial solvers using predefined affinity model~\cite{ChoECCV10,DuchennePAMI11, GoldPAMI96, LeeCVPR11}. The contributions of this paper are:
%(due to affinity model's limited capacity and disturbance)

i) We develop a novel supervised deep network based pipeline for graph matching, whereby the objective involves the permutation loss based on a Sinkhorn net rather than structured max-margin loss \cite{ChoICCV13} and pixel offset loss \cite{ZanfirCVPR18}. We argue that the permutation loss is a more inherent choice for the combinatorial nature graph matching (by relaxing it into linear assignment). Meanwhile, the permutation loss allows for the flexible handling of arbitrary number of nodes of graph for matching. In contrast, the number of nodes for matching in a graph is fixed and predefined in the problem structure in \cite{ChoICCV13}. To our best knowledge, this is the first time for adopting a permutation loss for learning graph matching -- a natural choice for its combinatorial nature.

ii) Our graph matching nets learn the node-wise feature (extracted from image in this paper) and the implicit structure information (including hyper-edge) by employing a graph convolutional network, together with the node-to-node cross-graph affinity function using additional layers. As such, the intra-graph information and cross-graph affinity are jointly learned given ground truth correspondence. Our network embeds both the node (image patch) feature and structure into the node-wise vector, and the node-to-node affinity layers are shared among all nodes. Such a design also allows for different numbers of nodes in different graph pairs for training and testing. To our best knowledge, this is the first time for adopting a graph neural network for learning graph matching (at least in computer vision).

iii) Experimental results including ablation studies show the effectiveness of our devised components including the permutation loss, the node-wise feature extract layer, graph convolutional network based node embedding, and the cross-graph affinity component. In particular, our method outperforms the deep learning peer method \cite{ZanfirCVPR18} in terms of matching accuracy. Our method also outperforms \cite{ChoICCV13} in accuracy while being more flexible as the method in \cite{ChoICCV13} requires constant number of nodes for matching in both training and testing sets. We also show the learning capability of our approach even when the training set and test set are from different object categories, which also outperforms~\cite{ZanfirCVPR18}.

%which limits their applicability to deal with similar graphs but with different number of nodes.

%Though there are active study on the above two-graph matching problem~\cite{GoldPAMI96,WykPAMI04,ChoECCV10}, the recent trend is (arguably) shifted to the setting for matching of multiple graphs. The main drivers may include: i) as more graphs are used, more information can be fused to boost the accuracy which otherwise can only be obtained by locally matching two graphs; ii) more and more practical problems call for the matching of more than two graphs, either in a offline batch mode or online incremental setting. In this paper, we focus on the multi-graph matching setting and start with the online scenario whereby the graph samples sequentially arrive and the solver aims to incrementally generating the matching results between any two graphs. In particular, we first consider and develop a dynamic clustering technique to group sequentially arrived graphs into multiple clusters, as such it is possible to perform matching in a distributed manner in each cluster.

%We have discussed many multi-graph matching methods however they only consider the offline setting i.e. all graphs are available for joint matching in one-shot. In applications graph data often arrives over time hence online incremental multi-graph matching is worth further study. One recent effort in this direction refers to \cite{chakraborty2016network}, however only first-order node-wise similarity is considered, and the proposed method is only applied on small-scale data.

\section{Related Work}
This paper focuses on learning of graph matching. Readers are referred to \cite{YanICMR16} for a comprehensive acquaintance. 
%As graph matching has been an active research area in computer vision, this paper only focuses on the most relevant ones on learning of graph matching. Readers are referred to a recent survey \cite{YanICMR16} for a comprehensive acquaintance. In addition, we also present related works from which the techniques used our method are derived.

\subsection{Modeling and Learning Affinity}
Recently a number of studies show various techniques for affinity function learning. Based on the extent to which ground truth correspondence information is used for training, methods are either unsupervised~\cite{LeordeanuIJCV12}, semi-supervised~\cite{LeordeanuICCV11}, or supervised~\cite{CaetanoPAMI09,ChoICCV13,ZanfirCVPR18}.

Previous graph matching affinity learning methods are mostly based on simple and shallow parametric models, which use popular distances (typically weighted Euclid distance) in the node and edge feature space plus a similarity kernel function (e.g. Gaussian kernel) to derive the final affinity score. In particular, a unified (shallow) parametric graph structure learning model is devised between two graphs in a vector form $\Phi(\mathcal{G}_1,\mathcal{G}_2,\pi)$ \cite{ChoICCV13}. 
%\begin{align}\notag
%\Phi=[\cdots,s_v(a_u,a_{\pi(u)}),\cdots,s_e(a_{uv},a_{\pi(u)\pi(v)})\cdots]^\top
%\end{align}
%By introducing weights on all elements of this feature map, one obtains a score function:
%$S(\mathcal{G}_i,\mathcal{G}_j,\pi,\beta)=\beta\Phi(\mathcal{G}_i,\mathcal{G}_j,\pi)$ where $\beta$ is a weight vector of $s_v$ and $s_u$.
The authors in \cite{ChoICCV13} observe that the above simple model can incorporate most previous shallow learning models, including \cite{CaetanoPAMI09,LeordeanuIJCV12,TorresaniECCV08}. Therefore, this method will be compared in our experiment.

%For instance, \cite{CaetanoPAMI09} adopts a 60-dimensional similarity function $s_v$ for feature points and a binary similarity $s_e$ for edges. \cite{LeordeanuIJCV12} uses a multi-dimensional $s_e$ to measure similarity without accounting for $s_v$. \cite{TorresaniECCV08} employs 2-dimensional $s_v$ and $s_e$ functions to model appearance similarity, occlusion, and geometric compatibility.

There is a seminal work \cite{ZanfirCVPR18} presenting a method adopting deep neural networks for learning the affinity matrix for graph matching. However, in Sec.~\ref{sec:further_discussion} we show that their pixel offset based loss function does not fit well with the combinatorial nature of graph matching. In addition, node embedding is not considered which is able to effectively capture the local structure of the node, which can go beyond second-order for more effective affinity modeling.
 
%By the criterion if deep networks are adopted or not, methods on affinity function learning can be generally divided into two threads. Most works belong to the first category that seek to learn a relatively simple (shallow) model with parameters of the node-to-node and edge-to-edge affinity, whereby
%\subsection{Learning of graph matching}
\subsection{Graph Neural Networks and Embedding}
Deep neural networks have been proven effective on spatial and sequential data, with CNN and RNN respectively.  Recently, there emerges a number of techniques for extracting high-order node embedding via deep networks, whose input i.e. graph is non-Euclidean data. Specifically, graph neural networks~(GNN)~\cite{ScarselliNN09} have been proposed whereby node features are aggregated from adjacent neighbors and different nodes can share the same transfer function. The output of GNN is invariant to permutations of graph elements. Many variants of GNN have been developed since~\cite{ScarselliNN09}, which is comprehensively discussed in~\cite{ZhouArxiv18}. In particular,  the SNDE model \cite{SDNE} is developed for deep node embedding by exploiting the first-order and second-order proximity jointly. Differing from the above deep embedding models, there are some shallow embedding models which are scalable on large networks including DeepWalk~\cite{deepwalk} based on random walk and node2vec~\cite{node2vec} inspired by skip-gram language model~\cite{Mikolov2013Efficient}. In particular, LINE~\cite{tang2015line} explicitly defines \emph{first-order} proximity and \emph{second-order} proximity and builds heuristics models for the two proximities. However, these methods, including the SNDE model cannot be used for end-to-end learning for graph matching. For this reason, we adopt the graph convolutional network~(GCN)~\cite{KipfICLR17} modeling graph structure whose parameters are learnable in an end-to-end fashion. 

%Neural network on graphs have shown great potential on computational-intensive problems like graph matching. For instance, \cite{NowakArxiv18} achieved comparative result on traveling salesman problem and graph matching with GNN. 

%Another recent work~\cite{BattagliaArxiv2018} unifies existing deep graph models into a framework named Graph Network~(GN). 

\subsection{Learning of Combinatorial Optimization}
Graph matching bears the combinatorial nature. There is an emerging thread using learning to seek efficient solution, especially with deep networks. In \cite{huang2019coloring}, the well known NP-hard problem for coloring very large graphs is addressed using deep reinforcement learning. The resulting algorithm can learn new state
of the art heuristics for graph coloring. While the Travelling Salesman Problem (TSP) is studied in \cite{kool2018attention} and the authors propose a graph attention network based method which learns a heuristic algorithm that employs neural network policy to find a tour. Deep learning for node set is also explored in \cite{zaheer2017deep} which seeks permutation invariant objective functions to a set of nodes.

In particular, \cite{nowak2018revised} shows a network based approach for solving the quadratic assignment problem. Their work focuses on learning the solver given previous defined affinity matrix. In contrast, this paper presents an end-to-end learning pipeline for learning the affinity function. In this sense, the two methods can be further integrated for practical applications. Moreover, for the less challenging linear assignment problem, which in fact can be solved with polynomial complexity e.g. the Hungarian algorithm~\cite{Kuhn1955Hungarian}, there also exist recently proposed network based new methods. The Sinkhorn Network~\cite{AdamsArxiv11} is developed for linear assignment learning in the sense of linear assignment given predefined assignment cost, which is designated to enforce doubly-stochastic regulation on any non-negative square matrix. It has been shown that Sinkhorn algorithm~\cite{sinkhorn1964relationship} is the approximate and differentiable version of Hungarian algorithm~\cite{MenaNIPS17}. More recently, the Sinkhorn AutoEncoder is proposed in \cite{PatriniArxiv18sinkhornAE} to minimize Wasserstein distance in AutoEncoders, and the work \cite{EmamiArxiv18} adopts reinforcement learning for learning a linear assignment solver. The Sinkhorn layer is also adopted on top of a deep convolutional network in DeepPermNet~\cite{SantaPAMI18}, which solves a permutation prediction problem. However, DeepPermNet is not invariant to input permutations and need a predefined node permutation as reference, thus it is unstable for two graph matching.

In comparison, our model consists of an affinity learning component which encodes the structure affinity into node-wise embeddings. As such, graph matching is relaxed into linear assignment solved by the Sinkhorn layer, which is also sometimes called permutation learning in literature.
\begin{table}[tb!]
    \centering
    \caption{Symbol notations. Subscript $s$ indexes image/graph.}
    \resizebox{0.48\textwidth}{!}
    {
    \begin{tabular}{r|l}
        \toprule
      %  \textbf{symbol} & \textbf{explanation} \\
    %    \midrule
        $I_s$ & input image $s$ \\
        $N$ & number of nodes in one graph \\
        $\mathbf{A}_s$ & adjacency matrix of graph $s$ \\
        $\mathcal{V}_s$ & vertex set of graph $s$ \\
        $\mathcal{E}_s$ & edge set of graph $s$ \\
        $P_{si}$ & coordinate of keypoint $i$ in image $s$ \\ \midrule
        $\mathbf{h}^{(k)}_{si}$ & feature vector of keypoint $i$, layer $k$ in graph $s$ \\
        $\mathbf{m}^{(k)}_{si}$ & message vector of keypoint $i$, layer $k$ in graph $s$\\
        $\mathbf{n}^{(k)}_{si}$ & node feature of keypoint $i$, layer $k$ in graph $s$ \\
        $\mathbf{M}^{(k)}$ & affinity matrix on $k$-th Sinkhorn iteration \\
        $\mathbf{S}$ & $N \times N$ matrix representing permutation \\ \bottomrule
    \end{tabular}
    }
    \vspace{-5pt}
    \label{tab:notation}
\end{table}

\section{Proposed Approach}
%\subsection{Preliminaries}
%\subsubsection{Notations}
% Bold lower class letters usually represent vectors, bold upper class letters usually represent matrix or higher-order tensors. Superscript usually stands for iteration number, and subscript usually stands for the index. 

%\subsubsection{Permutation and Doubly-Stochastic Matrix}
%\label{sec:PM_and_DSM}

%Permutation matrix and doubly-stochastic matrix are two concepts widely adopted in this paper.
%\begin{define}
%A permutation matrix is a square matrix whose elements are all zeros and ones. All its columns and rows sum to one.
%\end{define}
%i.e., for a permutation matrix $\mathbf{X}\in \{0, 1\}^{N\times N}$, there are 
%$$\begin{cases}
%    \mathbf{X} \mathbf{1} = \mathbf{1} \\
%    \mathbf{1}^\top \mathbf{X} = \mathbf{1}^\top
%\end{cases}$$

%\begin{define}
%A doubly-stochastic matrix is a square matrix with non-negative elements. All its columns and rows sum to one.
%\end{define}
%i.e., for a doubly-stochastic matrix  $\mathbf{X}\in \mathbb{R}^{N\times N}$, there are
%$$\begin{cases}
%    \mathbf{X} \mathbf{1} = \mathbf{1} \\
%    \mathbf{1}^\top \mathbf{X} = \mathbf{1}^\top
%\end{cases}$$

%The doubly-stochastic matrix is regarded as a convex relaxation of permutation matrix. A doubly-stochastic matrix can be restricted to a permutation matrix by solving a Linear Assignment Problem~(LAP) via Hungarian algorithm~\cite{Kuhn1955Hungarian}.

%\subsection{The Proposed Models}

%\subsubsection{Overview}
We present two models for matching $\mathcal{G}_1=(\mathcal{V}_1, \mathcal{E}_1)$ and $\mathcal{G}_2=(\mathcal{V}_2, \mathcal{E}_2)$: i) \underline{\textbf{p}}ermutation loss and \textbf{\underline{i}}ntra-graph \textbf{\underline{a}}ffinity based \textbf{\underline{g}}raph \textbf{\underline{m}}atching~(\textbf{PIA-GM}) and ii) \textbf{\underline{p}}ermutation loss and \textbf{\underline{c}}ross-graph \textbf{\underline{a}}ffinity based one~(\textbf{PCA-GM}). Both models are built upon a deep network which exploits both image feature and structure jointly, and a Sinkhorn network enabling differentiable permutation prediction and loss backpropagation. PCA-GM adopts an extra cross-graph component which aggregates cross-graph features, while PIA-GM only embeds intra-graph features. Fig.~\ref{fig:overview} summarizes both PIA-GM and PCA-GM. Symbols are shown in Tab. \ref{tab:notation}.

The proposed two models consist of a CNN image feature extractor, a graph embedding component, an affinity metric function and a permutation prediction component. Image features are extracted by CNN (VGG16 in the paper) as graph nodes, and aggregated through (cross-graph) node embedding component. The networks predict a permutation for node-to-node correspondence from raw pixel inputs. 
%We present the details of each component as follows.

\subsection{Feature Extraction}
We adopt a CNN for keypoints feature extraction, which are constructed by interpolating on CNN's feature map. For image $I_s$, the extracted feature on the keypoint $P_{si}$ is:
\begin{equation}
\label{eq:cnn}
    \mathbf{h}^{(0)}_{si} = \text{Interp}(P_{si}, \text{CNN}(I_s))
\end{equation}
where $\text{Interp}(P, X)$ interpolates on point $P$ from tensor $X$ via bilinear interpolation. $\text{CNN}(I)$ performs CNN on image $I$ and outputs a feature tensor. Taking the idea of Siamese Network \cite{BromleyNIPS94}, two input images share the same CNN structure and weights. To fuse both local structure and global semantic information, feature vectors from different layers of CNN are extracted. We choose VGG16 pretrained with ImageNet~\cite{deng2009imagenet} as the CNN embodiment in line with \cite{ZanfirCVPR18}.

%Note the CNN can take any form of backbone structure and can be initialized with pretrained weights e.g. on ImageNet~\cite{deng2009imagenet}.

\subsection{Intra-graph Node Embedding}
It has been shown that methods exploiting graph structure can produce robust matching~\cite{YanICMR16}, compared to point based methods~\cite{FischlerCACM81,ZhangIJCV94}. In PIA-GM, graph affinity is constructed by a multi-layer embedding component which models the higher-order information. The message passing scheme is inspired by GCN~\cite{KipfICLR17}, where features are effectively aggregated from adjacency nodes, and the node itself:
\begin{align}
\label{eq:gnn_msg}
    \mathbf{m}^{(k)}_{si} =& \frac{1}{|(i, j) \in \mathcal{E}_s|}\sum_{j: (i, j) \in \mathcal{E}_s} f_{\textit{msg}}(\mathbf{h}^{(k-1)}_{sj})\\
\label{eq:gnn_node}
    \mathbf{n}^{(k)}_{si} =& f_{\textit{node}}(\mathbf{h}^{(k-1)}_{si})\\
\label{eq:gnn_update}
    \mathbf{h}^{(k)}_{si} =& f_{\textit{update}}(\mathbf{m}^{(k)}_{si}, \mathbf{n}^{(k)}_{si})
\end{align}

Eq.~(\ref{eq:gnn_msg}) is the message passing along edges and $f_{\textit{msg}}$ is the message passing function. The aggregated features from adjacent nodes are normalized by the total number of adjacent nodes, as a common practice in GCN, in order to avoid the bias due to the different numbers of neighbors owned by different nodes. Eq.~(\ref{eq:gnn_node}) is the message passing function for each node and it contains a node's self-passing function $f_{\textit{node}}$. With $f_{\textit{update}}$, Eq.~(\ref{eq:gnn_update}) accumulates information to update the state of node $i$, and $f_{\textit{msg}}, f_{\textit{node}}, f_{\textit{update}}$ may take any differentiable mapping from vector to vector. Here we implement $f_{\textit{msg}}, f_{\textit{node}}$ as neural networks with ReLU activation, and $f_{\textit{update}}$ is a summation function. We denote Eq.~(\ref{eq:gnn_update}) as graph convolution (GConv) between layer $k-1$ and $k$:
\begin{equation}
\label{eq:gconv}
    \{\mathbf{h}_{si}^{(k)}\} = \text{GConv}(\mathbf{A}_s, \{\mathbf{h}_{si}^{(k-1)}\}), \quad i \in \mathcal{V}_s
\end{equation}
which denotes a layer of our node embedding net. Message passing paths are encoded by adjacency matrix $\mathbf{A}_s \in \{0, 1\}^{N\times N}$. Note that $\mathbf{h}_i^{(0)}$ is the CNN feature of node $i$.

\subsection{Cross-graph Node Embedding} 
%PCA is built upon PIA, with an extra cross-graph matching layer that models cross-graph affinity. Both PCA and PIA share the same structure of CNN module, affinity composition and Sinkhorn layer. PCA owns a GNN module which allows cross-graph message passing, while PIA's GNN is only capable to embed intra-graph features. Details will be given on the cross-graph GNN module in the following context. Details on the cross-graph GNN module are given in Algorithm~\ref{alg:cross-graph}.

%The intra-graph aggregation step in PCA is similar to the GNN module in PIA. Recall the graph convolution operator introduced in Eq.~(\ref{eq:gconv}). Several layers of graph convolution is adopted to aggregate intra-graph information, before applying the cross-graph matching module. After the cross-graph matching layer, another several layers of intra-graph convolution are adopted.

\setlength{\textfloatsep}{5pt}
\begin{algorithm}[t]
{\small{
    \label{alg:cross}
    \caption{\textbf{Cross-graph node embedding}}
    \KwIn{$(k-1)$-th layer features $\{\mathbf{h}_{1i}^{(k-1)},\mathbf{h}_{2j}^{(k-1)}\}_{i\in \mathcal{V}_1, j\in \mathcal{V}_2}$}
    // similarity prediction Eq.~(\ref{eq:aff}, \ref{eq:sinkhorn})\\
	build $\hat{\mathbf{M}}$ from $\{\mathbf{h}_{1i}^{(k-1)},\mathbf{h}_{2j}^{(k-1)}\}$ by Eq.~(\ref{eq:aff});\\
	$\hat{\mathbf{S}} \leftarrow \text{Sinkhorn}(\hat{\mathbf{M}})$; \\
	// cross-graph aggregation Eq.~(\ref{eq:cross_gnn_msg}, \ref{eq:cross_gnn_node}, \ref{eq:cross_gnn_update})\\
	$\{\mathbf{h}_{1i}^{(k)}\} \leftarrow \text{CrossConv}(\hat{\mathbf{S}}, \{\mathbf{h}_{1i}^{(k-1)}\}_{i\in \mathcal{V}_1}, \{\mathbf{h}_{2j}^{(k-1)}\}_{j\in \mathcal{V}_2})$; \\
	$\{\mathbf{h}_{2j}^{(k)}\} \leftarrow \text{CrossConv}(\hat{\mathbf{S}}^{\top}, \{\mathbf{h}_{2j}^{(k-1)}\}_{j\in \mathcal{V}_2}, \{\mathbf{h}_{1i}^{(k-1)}\}_{i\in \mathcal{V}_1})$; \\
	\KwOut{$k$-th layer features $\{\mathbf{h}_{1i}^{(k)},\mathbf{h}_{2j}^{(k)}\}_{i\in \mathcal{V}_1, j\in \mathcal{V}_2}$}
}}
\end{algorithm}
We explore improvement over intra-graph embedding by a cross-graph aggregation step, whereby features are aggregated from nodes with similar features in the other graph. First, we utilize graph affinity features from shallower embedding layers to predict a doubly-stochastic similarity matrix (see details in Sec.~\ref{subsec:sinkhorn}). The predicted similarity matrix $\hat{\mathbf{S}}$ encodes the similarity among nodes of two graphs. The message passing scheme is similar to intra-graph convolution in Eq.~(\ref{eq:gconv}), with adjacency matrix replaced by $\hat{\mathbf{S}}$, and features are aggregated from the other graph. In our experiments, we will show this simple scheme works more effectively than a more complex iterative procedure.
\begin{align}
\label{eq:cross_gnn_msg}
    \mathbf{m}^{(k)}_{1i} =& \sum_{j \in \mathcal{V}_2} \hat{\mathbf{S}}_{i, j} f_{\textit{msg-cross}}(\mathbf{h}^{(k-1)}_{2j})\\
\label{eq:cross_gnn_node}
    \mathbf{n}^{(k)}_{1i} =& f_{\textit{node-cross}}(\mathbf{h}^{(k-1)}_{1i})\\
\label{eq:cross_gnn_update}
    \mathbf{h}^{(k)}_{1i} =& f_{\textit{update-cross}}(\mathbf{m}^{(k)}_{1i}, \mathbf{n}^{(k)}_{1i})
\end{align}
where $f_{\textit{msg-cross}}, f_{\textit{node-cross}}$ are taken as identity mapping, $f_{\textit{update-cross}}$ is a concatenation of two input feature tensors, followed by a fully-connected layer. For pair of graphs $\mathcal{G}_1=(\mathcal{V}_1, \mathcal{E}_1), \mathcal{G}_2=(\mathcal{V}_2, \mathcal{E}_2)$, the cross-graph aggregation scheme is summarized by CrossConv$(\cdot)$ in Alg.~\ref{alg:cross}, where $\hat{\mathbf{S}}$ denotes the predicted correspondence from $\mathcal{G}_2$ to $\mathcal{G}_1$ and $\hat{\mathbf{S}}^\top$ denotes such relation from $\mathcal{G}_1$ to $\mathcal{G}_2$.
%Eq.~(\ref{eq:cross_gnn_msg}, \ref{eq:cross_gnn_node}, \ref{eq:cross_gnn_update}) is denoted as
\if0
\begin{equation}
    \label{eq:cross_gconv}
\begin{split}
    \{\mathbf{h}_{1i}^{(k+1)}\}_{i\in \mathcal{V}_1} & = \text{CrossConv}(\hat{\mathbf{S}}, \{\mathbf{h}_{1i}^{(k)}\}_{i\in \mathcal{V}_1}, \{\mathbf{h}_{2j}^{(k)}\}_{j\in \mathcal{V}_2}) \\
    \{\mathbf{h}_{2j}^{(k+1)}\}_{j\in \mathcal{V}_2} & = \text{CrossConv}(\hat{\mathbf{S}}^\top, \{\mathbf{h}_{2j}^{(k)}\}_{j\in \mathcal{V}_2}, \{\mathbf{h}_{1i}^{(k)}\}_{i\in \mathcal{V}_1})
\end{split}
\end{equation}
where $\hat{\mathbf{S}}$ denotes the predicted correspondence from $\mathcal{G}_2$ to $\mathcal{G}_1$ and $\hat{\mathbf{S}}^\top$ denotes such relation from $\mathcal{G}_1$ to $\mathcal{G}_2$. 
\fi

\subsection{Affinity Metric Learning} 
By using the above embedding model, the structure affinity between two graphs have been encoded into the node-to-node affinity in the embedding space. As such, it allows for reducing the traditional second-order affinity matrix $\mathbf{K}$ in Eq.~(\ref{eq:lawler_qap}) into a linear one. Let $\mathbf{h}_{1i}$ be feature $i$ from first graph, $\mathbf{h}_{2j}$ be feature $j$ from the other graph:
\begin{equation}
    \mathbf{M}^{(0)}_{i,j} = f_{\textit{aff}}(\mathbf{h}_{1i}, \mathbf{h}_{2j}), \quad i \in \mathcal{V}_1, j \in \mathcal{V}_2
\end{equation}

The affinity matrix $\mathbf{M}^{(0)} \in {\mathbb{R}^+}^{N\times N}$ contains the affinity score between two graphs. $\mathbf{M}^{(0)}_{i,j}$ means the similarity between node $i$ in the first graph and node $j$ in the second, considering the higher-order information in graphs.

One can set $f_{\textit{aff}}$ a bi-linear mapping followed by an exponential function which ensures  all elements are positive\footnote{We have also tried other more flexible fully-connected layers, while we find the exponential function is simple and more stable for learning.}.
\begin{equation}
\label{eq:aff}
    \mathbf{M}^{(0)}_{i,j} = \text{exp}\left(\frac{\mathbf{h}_{1i}^\top \mathbf{A} \mathbf{h}_{2j}}{\tau}\right)
\end{equation}

Consider the feature vectors have $m$ dimensions, i.e. $\forall i \in \mathcal{V}_1, j \in \mathcal{V}_2, \mathbf{h}_{1i}, \mathbf{h}_{2j} \in \mathbb{R}^{m\times 1}$. $\mathbf{A} \in \mathbb{R}^{m \times m}$ contains learnable weights of this affinity function. $\tau$ is a hyper parameter for numerical concerns. For $\tau>0$, with $\tau \rightarrow 0^+$, Eq.~(\ref{eq:aff}) becomes more discriminative.

\subsection{Sinkhorn Layer for Linear Assignment} \label{subsec:sinkhorn}
Given the linear assignment affinity matrix in Eq.~(\ref{eq:aff}), we adopt Sinkhorn for the linear assignment task. Sinkhorn operation takes any non-negative square matrix and outputs a doubly-stochastic matrix, which is a relaxation of the permutation matrix. This technique has been shown effective for network based permutation prediction~\cite{AdamsArxiv11,SantaPAMI18}. For $\mathbf{M}^{(k-1)} \in {\mathbb{R}^+}^{N\times N}$, the Sinkhorn operator is
\begin{align}
\label{eq:sk_row}
    \mathbf{M}^{(k)\prime} =& \mathbf{M}^{(k-1)} \oslash (\mathbf{M}^{(k-1)}\mathbf{1} \mathbf{1}^\top)\\
\label{eq:sk_col}
    \mathbf{M}^{(k)} =& \mathbf{M}^{(k)\prime} \oslash ( \mathbf{1} \mathbf{1}^\top \mathbf{M}^{(k)\prime})
\end{align}
$\oslash$ means element-wise division, and $\mathbf{1} \in \mathbb{R}^{N\times 1}$ is a column vector whose elements are all ones. Sinkhorn algorithm works iteratively by taking row-normalization of Eq.~(\ref{eq:sk_row}) and column-normalization of Eq.~(\ref{eq:sk_col}) alternatively.

By iterating Eq.~(\ref{eq:sk_row}, \ref{eq:sk_col}) until convergence, we get a doubly-stochastic matrix. This doubly-stochastic matrix $\mathbf{S}$ is treated as our model's prediction in training. 
\begin{equation}
\label{eq:sinkhorn}
    \mathbf{S} = \text{Sinkhorn}(\mathbf{M}^{(0)}) 
\end{equation}

For testing, Hungarian algorithm~\cite{Kuhn1955Hungarian} is performed on $\mathbf{S}$ as a post processing step to discretize output into a permutation matrix. Sinkhorn operation is fully differentiable because only matrix multiplication and element-wise division are taken. It can be efficiently implemented with the help of PyTorch's automatic differentiation feature~\cite{serratosa2011automatic}.
 
\subsection{Permutation Cross-Entropy Loss}
Our methods directly utilize ground truth node-to-node correspondence, i.e. permutation matrix, as the supervised information for end-to-end training. Since Sinkhorn layer in Eq.~(\ref{eq:sinkhorn}) is capable to transform any non-negative matrix into doubly-stochastic matrix, we propose a linear assignment based permutation loss that evaluates the difference between predicted doubly-stochastic matrix and ground truth permutation matrix for training.

Cross entropy loss is adopted to train our model end-to-end. We take the ground truth permutation matrix $\mathbf{S}^{gt}$, and compute the cross entropy loss between $\mathbf{S}$ and $\mathbf{S}^{gt}$. It is denoted as permutation loss, and this is the main method adopted to train our deep graph matching model $L_{\textit{perm}}$:
\begin{equation}
\label{eq:perm_loss}
    - \sum_{i \in \mathcal{V}_1, j \in \mathcal{V}_2} \left(\mathbf{S}^{gt}_{i,j} \log \mathbf{S}_{i,j} + (1-\mathbf{S}^{gt}_{i,j}) \log (1-\mathbf{S}_{i,j}) \right)
\end{equation}

Note the competing method GMN~\cite{ZanfirCVPR18} applies a pixel offset based loss namely ``displacement loss''. Specifically it computes an offset vector $\mathbf{d}$ by a weighted sum of all matching candidates. The loss is given as the difference between predicted location and ground truth location.
\begin{align}
    \mathbf{d}_i =& \sum_{j \in V_2} \left( \mathbf{S}_{i, j} P_{2j} \right)- P_{1i}\\
\label{eq:offset_loss}
    L_{\textit{off}} =& \sum_{i \in V_1} \sqrt{||\mathbf{d}_i - \mathbf{d}^{gt}_i||^2 + \epsilon}
\end{align}
where $\{P_{1i}\}, \{P_{2j}\}$ are the coordinates of keypoints in first and second image, respectively. While $\epsilon$ is a small value ensuring numerical robustness. In comparison, our cross entropy loss can directly learn a linear assignment cost based permutation loss in an end-to-end fashion.

\subsection{Further Discussion}
\label{sec:further_discussion}
\noindent \textbf{Pairwise affinity matrix vs. embedding}.
Existing graph matching methods focus on modeling second-order~\cite{ChoECCV10,LeordeanuICCV05} and higher-order~\cite{LeeCVPR11,YanCVPR15} feature with an explicitly pre-defined affinity matrix or tensor. The affinity information can be encoded in an $N^2 \times N^2$ affinity matrix. Optimization techniques are applied to maximize graph affinity.

In contrast, we resort to the node embedding technique with two merits. First, the space complexity can be reduced to $N \times N$. Second, the pairwise affinity matrix $\mathbf{K}$ in Eq.~(\ref{eq:lawler_qap}) can only encode the edge information, while the embedding model can implicitly encode higher-order information. 

\noindent \textbf{Sinkhorn net vs. spectral matching}.
GMN~\cite{ZanfirCVPR18} adopts spectral matching (SM)~\cite{LeordeanuICCV05} which is differentiable for back propagation. While we adopt the Sinkhorn net instead. In fact, the input of Sinkhorn is of complexity $O(N^2)$ while it is $O(N^4)$ for spectral matching. However, in SM we observe more iterations to convergence. Such iteration may bring negative effect to gradient's back propagation. In fact, spectral matching is for graph matching while Skinhorn net is for linear assignment, which is relaxed from the graph matching task by our embedding component.

%Experiments show the superiority of GNN over SM in linear assignment learning problems.

\setlength{\textfloatsep}{20pt}
\begin{figure}[!tb]
    \centering
    \includegraphics[width=0.42\textwidth]{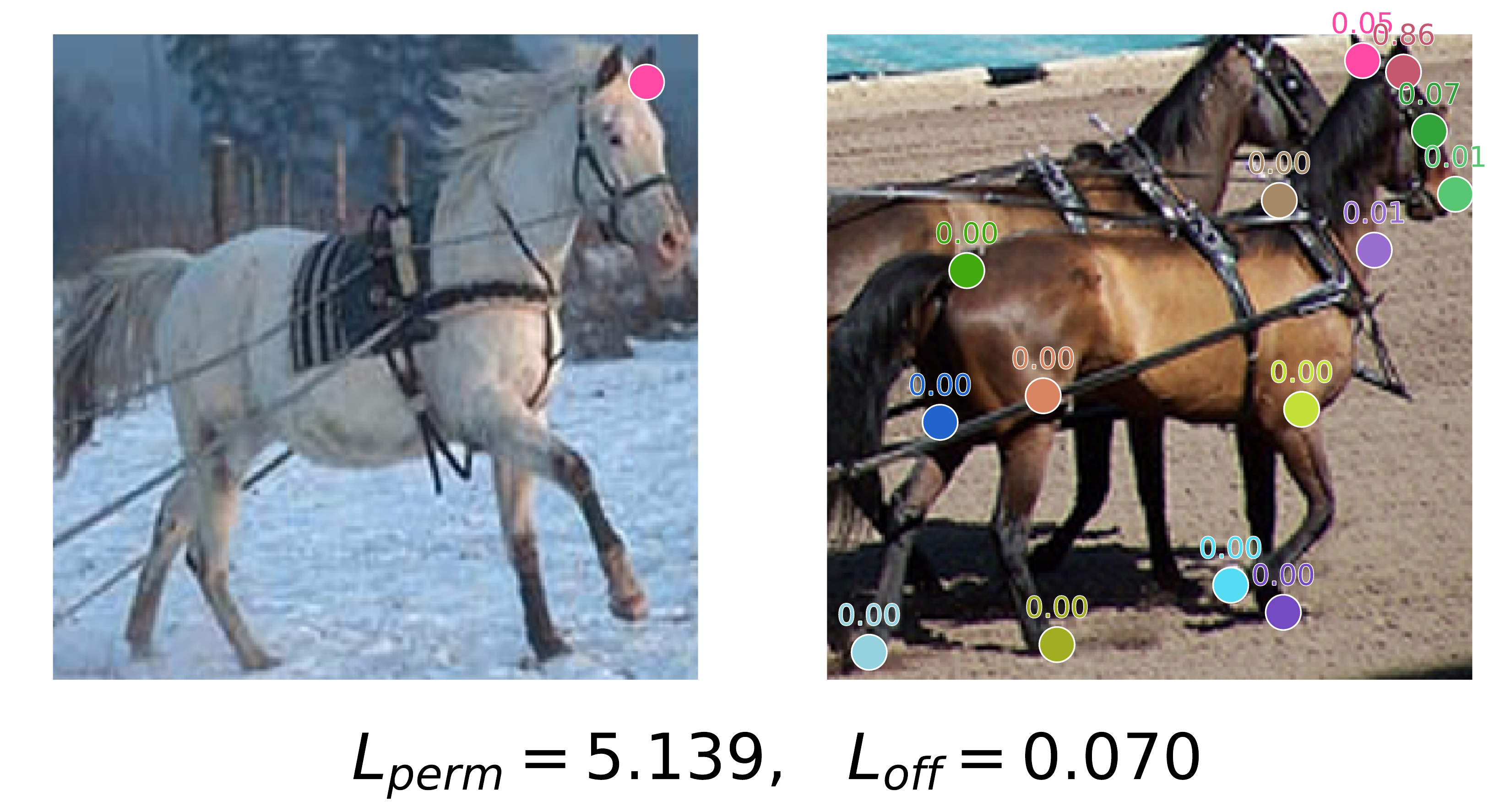}
    \vspace{-5pt}
    \caption{Failure case of offset loss: source image (left) and target image (right) with matching candidates, where numbers denote the probability of predicted matching. Ground truth matching nodes are colored in rose (only receives 0.05 probability by this poor prediction). Offset loss is computed by a weighted sum among all candidates, resulting in a misleading low loss 0.070. In this case offset loss fails to provide supervision on distinguishing left/right ears. Our permutation loss, on the contrary, issues a reasonably high loss 5.139.}
    \label{fig:loss}
\vspace{-10pt}
\end{figure}

\noindent \textbf{Pixel offset loss vs. permutation loss}. The loss function adopted by GMN~\cite{ZanfirCVPR18} is an offset loss named ``displacement loss''. The loss takes the weighted sum of all candidate points, and compute the offset vector from the original image to the source image. In training, GMN tries to minimize the variance between predicted offset vector and ground truth offset vector. In comparison, with the help of Sinkhorn net, we adopt a combinatorial permutation loss which is computed as the cross entropy between predicted result and ground truth permutation. Such permutation loss takes the ground truth permutation directly as supervision, and utilize such information for end-to-end training. 

Fig.~\ref{fig:loss} gives an example for the failure case of offset loss. In this case, the offset loss is unreasonably low, but the permutation loss provides correct information. Experiments also show that models trained with our permutation loss exceed offset loss models in matching accuracy.

\section{Experiments}
\subsection{Metrics and Peer Methods}
We evaluate the matching accuracy between two given graphs. In the evaluation period, two graphs are given with same number of nodes $N$. Each node in one graph is labeled to another node in the other graph. The model predicts a correspondence between two graphs. Such correspondence is represented by a permutation matrix. %The matching accuracy is computed from the predicted permutation matrix and ground truth permutation matrix.

\begin{comment}
First we define a logical $\text{AND}$ function by:
\begin{equation}
    \text{AND}(a, b) = \left\{
\begin{aligned}
1 & , & a=b=1 \\
0 & , & \text{otherwise}
\end{aligned}
\right.
\end{equation}
\end{comment}

The matching accuracy is computed from the permutation matrix, by the number of correctly matched keypoint pairs averaged by the total number of keypoint pairs. For a predicted permutation matrix $\mathbf{S}^{pred} \in \{0, 1\}^{N\times N}$ and a ground truth permutation $\mathbf{S}^{gt} \in \{0, 1\}^{N\times N}$, matching accuracy is computed by
\begin{equation}
    \text{acc} = \sum \text{AND}(\mathbf{S}^{pred}_{i, j}, \mathbf{S}^{gt}_{i, j}) / N
    %\text{acc} = \frac{|P^{pred}\bigcap P^{gt}|}{|P^{gt}|}
\end{equation}
%\begin{equation}
%    \text{acc} = \{\# \text{ of } \mathbf{S}^{pred}_{i,j}=\mathbf{S}^{gt}_{i, j}=1\} / {N}
%\end{equation}
where $\text{AND}$ is the logical function.

The evaluation involves the following peer methods:

\textbf{GMN.} Graph Matching Network~(GMN) is the seminal model proposed in~\cite{ZanfirCVPR18}. GMN adopts VGG16~\cite{simonyanICLR14vgg} network to extract image features. First-order and second-order features are extracted from shallower layer (relu4\_2) and deeper layer (relu5\_1) of VGG16, respectively. GMN models graph matching affinity via an unlearnable graph matching solver namely spectral matching~(SM)~\cite{LeordeanuICCV05}. This model is class-agnostic, meaning it learns an universal model for all instance classes. Two graphs are constructed by Delaunay triangulation and fully-connected topology, respectively. GMN is the first end-to-end deep learning method for graph matching. Note the major difference is that the loss function is an offset based loss by Eq.~(\ref{eq:offset_loss}). We follow~\cite{ZanfirCVPR18} and re-implement GMN with PyTorch as the source code is not publicly available. 

\textbf{HARG-SSVM.} This is the structured SVM based learning graph matching method  \cite{ChoICCV13}, as a baseline of learning graph matching without deep learning. HARG-SSVM is a class-specific method, where graph models are learned for each class. We use the source code released by the authors upon their approval. The original setting in \cite{ChoICCV13} assumes that the keypoints of the object to be matched is unknown, and the keypoint candidates are proposed by Hessian detector~\cite{mikolajczykIJCV04hessian}. In our setting, however, all candidate keypoints are known to the model. Therefore, we slightly modify the original code. From all candidate points found by the Hessian detector, we assign the nearest neighbor from ground truth point as matching candidate. This practice is originally taken in the training process of HARG-SSVM. Graphs are created with hand-crafted edge features named HARG.

\textbf{PIA-GM/PCA-GM.} Our methods adopt VGG16~\cite{simonyanICLR14vgg} as backbone CNN, and extract features from relu4\_2 and relu5\_1 for fair comparison with \cite{ZanfirCVPR18}. These two feature vectors are concatenated to fusion both local and global features. In PIA-GM, affinity is modeled by a 3 intra-embedding layers while in PCA-GM it is a stack of 1 intra layer, 1 cross layer and 1 intra layer, both followed by affinity mapping in Eq.~(\ref{eq:aff}). Each GNN layer has a feature dimension of 2048. Permutation loss in Eq.~(\ref{eq:perm_loss}) is used. Input graphs are both constructed by Delaunay triangulation and we empirically set $\tau=0.005$ in Eq.~(\ref{eq:aff}). Our models are implemented by PyTorch.

\textbf{GMN-PL \& PIA/PCA-GM-OL.} GMN-PL and PIA/PCA-GM-OL are variants from GMN~\cite{ZanfirCVPR18} and our proposed PIA/PCA-GM, respectively. GMN-PL changes the offset loss in GMN to permutation loss, with all other configurations unchanged. While PIA/PCA-GM-OL switch the permutation loss to offset loss, leaving all other components unchanged. 

For natural image experiments, we draw two images from the dataset, and build two graphs containing the same number of nodes. The graph structure is agnostic, and is constructed according to methods' configurations (see discussions above). The CNN weight is initialized by a pretrained model on ImageNet~\cite{deng2009imagenet} classification dataset. 

\begin{table*}[tb!]
	\centering
	%\vspace{8pt}
	\caption{Accuracy (\%) on Pascal VOC Keypoint. Note after replacing the offset loss by permutation loss, GMN-PL outperforms GMN~\cite{ZanfirCVPR18} almost in all categories. While our method PIA-GM's performance degenerates when its permutation loss is changed to offset loss.}
	\resizebox{1\textwidth}{!}
	{
		\begin{tabular}{r|cccccccccccccccccccc|c}
			\toprule%\multirow{2}{*}{Model}
			method&aero&bike&bird&boat&bottle&bus&car&cat&chair&cow&table&dog&horse&mbike&person&plant&sheep&sofa&train&tv&mean \\\midrule
			GMN & 31.9 & 47.2 & 51.9 & 40.8 & 68.7 & 72.2 & 53.6 & 52.8 & 34.6 & 48.6 & \textbf{72.3} & 47.7 & 54.8 & 51.0 & 38.6 & 75.1 & 49.5 & 45.0 & 83.0 & 86.3 & 55.3\\
            GMN-PL& 31.1 & 46.2 & 58.2 & 45.9 & 70.6 & 76.4 & 61.2 & 61.7 & \textbf{35.5} & 53.7 & 58.9 & 57.5 & 56.9 & 49.3 & 34.1 & 77.5 & 57.1 & 53.6 & 83.2 & 88.6 & 57.9\\
            PIA-GM-OL& 39.7 & \textbf{57.7} & 58.6 & 47.2 & 74.0 & 74.5 & 62.1 & 66.6 & 33.6 & 61.7 & 65.4 & 58.0 & 67.1 & 58.9 & 41.9 & 77.7 & 64.7 & 50.5 & 81.8 & 89.9 & 61.6\\
            PIA-GM& \textbf{41.5} & 55.8 & 60.9 & \textbf{51.9} & 75.0 & 75.8 & 59.6 & 65.2 & 33.3 & \textbf{65.9} & 62.8 & \textbf{62.7} & 67.7 & 62.1 & 42.9 & \textbf{80.2} & 64.3 & \textbf{59.5} & 82.7 & 90.1 & 63.0\\
            PCA-GM& 40.9 & 55.0 & \textbf{65.8} & 47.9 & \textbf{76.9} & \textbf{77.9} & \textbf{63.5} & \textbf{67.4} & 33.7 & 65.5 & 63.6 & 61.3 & \textbf{68.9} & \textbf{62.8} & \textbf{44.9} & 77.5 & \textbf{67.4} & 57.5 & \textbf{86.7} & \textbf{90.9} & \textbf{63.8} \\
			\bottomrule
	\end{tabular}}
	%\vspace{8pt}
    \label{tab:voc_main}
	\vspace{-10pt}
\end{table*}
\begin{figure}[!tb]
    \centering
    \includegraphics[width=0.48\textwidth]{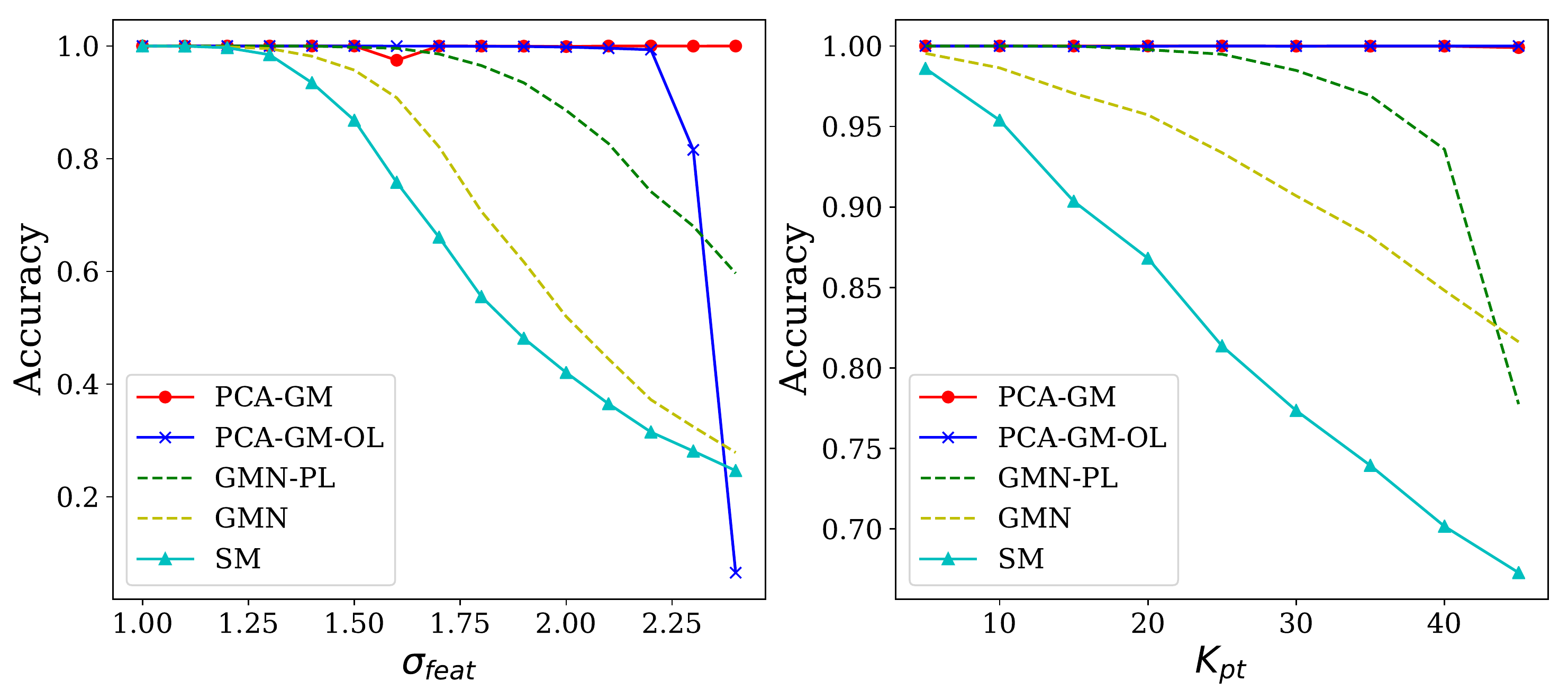}
    \vspace{-15pt}
    \caption{Synthetic test with noise on feature vectors of node and edge, and keypoint numbers on methods with different affinity models and losses. Default: $K_{\textit{pt}}=20, \sigma_{\textit{feat}}=1.5, \sigma_{\textit{coo}}=10$.}
    \label{fig:synthetic}
    \vspace{-10pt}
\end{figure}

\subsection{Synthetic Graphs}
Evaluation is first performed on synthetic graphs generated in line with the protocol in \cite{ChoECCV10}. Ground truth graphs are generated with a given number of keypoints $K_{\textit{pt}}$, each with a 1024-dimensional (512 for nodes and 512 for edges) random feature in $\mathcal{U}(-1, 1)$ (simulating CNN features), and a random 2d-coordinate in $\mathcal{U}(0, 256)$. During training and testing, we draw disturbed graphs, with Gaussian noise $\mathcal{N}(0, \sigma_{\textit{feat}}^2)$ added to features, and keypoint coordinates blurred by a random affine transform, plus another random noise of $\mathcal{N}(0, \sigma_{\textit{coo}}^2)$. Note that there is no CNN feature extractor adopted, only graph modeling approaches and loss metrics are compared. The matching accuracy of PCA-GM, PCA-GM-OL, GMN-PL, GMN and unlearning SM is evaluated with respect to $K_{\textit{pt}}$ and $\sigma_{\textit{feat}}$. For each trial, 10 different graphs are generated and accuracy is averaged. Experimental results in Fig.~\ref{fig:synthetic} show the robustness of PCA-GM against feature deformation and complicated graph structure.

\subsection{Pascal VOC Keypoints}
We perform experiments on Pascal VOC dataset~\cite{Everingham2010Pascal} with Berkeley annotations of keypoints~\cite{bourdev2009poselets_VOCkeypoint}. It contains 20 classes of instances with labeled keypoint locations. Following the practice of peer methods~\cite{ZanfirCVPR18}, the original dataset is filtered to 7,020 annotated images for training and 1,682 for testing. All instances are cropped around its bounding box and resized to $256\times 256$, before passed to the network. Pascal VOC Keypoint is a difficult dataset because instances may vary from its scale, pose and illumination, and the number of inliers ranges from $6$ to $23$.
%The average matching accuracy is the mean of 20 categories.

We test on Pascal VOC Keypoint~\cite{bourdev2009poselets_VOCkeypoint} and evaluate on 20 Pascal categories. We compare GMN, GMN-PL, PIA-GM-OL, PIA-GM, PCA-GM and give detailed experimental results in Tab.~\ref{tab:voc_main}. Our proposed models PIA-GM-OL, PIA-GM, PCA-GM outperform in most categories, including the mean accuracy over 20 categories. Our implementation of PCA-GM runs at $\sim 18$ pairs per second in training, on dual RTX2080Ti GPUs. The result shows the superiority of the linear assignment loss over offset loss in training, embedding and Sinkhorn over fixed SM~\cite{LeordeanuICCV05} in affinity modeling, and cross-graph embedding over intra-graph embedding.

\begin{table}[tb!]
\centering
\caption{Accuracy (\%) on Willow ObjectClass. GMN-VOC means model trained on Pascal VOC Keypoint, likewise for Willow.}
\resizebox{0.48\textwidth}{!}
	{
\begin{tabular}{r|ccccc}
\toprule
method & face & m-bike & car & duck & w-bottle   \\ \midrule 
HARG-SSVM~\cite{ChoICCV13} & 91.2 & 44.4 & 58.4 & 55.2 & 66.6 \\
GMN-VOC~\cite{ZanfirCVPR18} & 98.1 & 65.0 & 72.9 & 74.3 & 70.5 \\
GMN-Willow~\cite{ZanfirCVPR18} & 99.3 & 71.4 & 74.3 & 82.8 & 76.7 \\
PCA-GM-VOC & \textbf{100.0} & 69.8 & 78.6 & 82.4 & 95.1 \\
PCA-GM-Willow &\textbf{100.0} & \textbf{76.7} & \textbf{84.0} & \textbf{93.5} & \textbf{96.9} \\
\bottomrule
\end{tabular}
}
%\vspace{1pt}
\vspace{-10pt}
\label{tab:willow_main}
\end{table}

\subsection{Willow ObjectClass}
Willow ObjectClass dataset is collected by \cite{ChoICCV13} for real images. This dataset consists of 5 categories from Caltech-256 (face, duck and wine bottle) and Pascal VOC 2007 (car and motorbike), each with at least 40 images. Images are resized to $256\times 256$ if it is passed to CNN. This dataset is considered easier than Pascal VOC Keypoint, because all images inside the same category are aligned in their pose, and it lacks scale, background and illumination changes. 

We follow the protocol built by authors of~\cite{ChoICCV13} for fair evaluation. HARG-SSVM is trained and evaluated on this willow dataset. For other competing methods, we initialize their weights on Pascal VOC Keypoint dataset, with all VOC 2007 car and motorbike images removed. They are denoted as GMN-VOC and PCA-GM-VOC. They are later finetuned on the willow dataset as GMN-Willow and PCA-GM-Willow, reaching even higher result in evaluation.  Note that HARG-SSVM is a class-specific model, but GMN and PCA-GM are both class-agnostic. Tab.~\ref{tab:willow_main} shows our proposed PCA-GM almost surpasses all competing methods in all categories of Willow Object Calss dataset.

\begin{figure*}[!tb]
    \centering
    \includegraphics[width=0.95\textwidth]{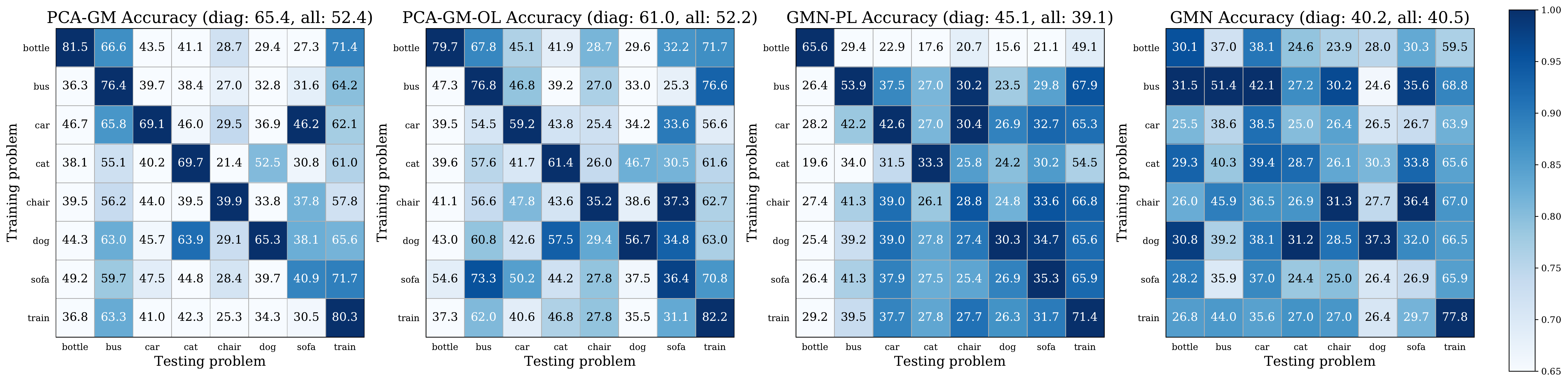}
        \vspace{-5pt}
    \caption{Confusion matrix of eight categories of objects from Pascal VOC Keypoint. Models are trained on categories on the $y$-axis, and testing results are given on categories on the $x$-axis. Note that accuracy does not degenerate much for PCA-GM between similar categories (such as cat and dog). Numbers in matrices are the corresponding accuracy and the color map stands for accuracy normalized by the highest accuracy on this category in current matrix. Note that the color filled in cells does not denote the absolute value of accuracy among different categories and matrices. Accuracy for elements in diagonal and overall for each confusion matrix are shown in bracket on the top of each matrix. We follow the train/test split provided by the benchmark for each category.}
    \vspace{-12pt}
    \label{fig:confusion_matrix}
\end{figure*}

\subsection{Further Study}
\begin{table}[tb!]
\centering
\caption{Ablation study on proposed components on Pascal VOC Keypoint. Tick denotes the learning is activated for the column. For VGG16 feature it means it is fine-tuned using the graph matching training data, otherwise the pretrained VGG16 via ImageNet.}
\resizebox{0.48\textwidth}{!}
	{
\begin{tabular}{cccc|c}
\toprule
 VGG16 &  intra-graph  &  cross-graph  & affinity & \multirow{2}{*}{accuracy}  \\  
  feature &  embedding &  embedding & metric &   \\ \midrule 
\checkmark & \checkmark & \checkmark & \checkmark & \textbf{63.8}  \\
\checkmark & \checkmark & \checkmark & $\times$ & 63.6  \\
\checkmark & \checkmark & $\times$ & $\times$ & 62.1   \\
\checkmark & $\times$ & $\times$ & $\times$ & 54.8  \\
$\times$ & $\times$ & $\times$ & $\times$ & 41.9 \\
\bottomrule
\end{tabular}
}
\vspace{-10pt}
\label{tab:voc_modules}
\end{table}

\textbf{PCA-GM components}. Ablation study with different PCA-GM components trained/untrained is reported in Tab.~\ref{tab:voc_modules}. It shows the usefulness of all our components. VGG16 is initialized with pretrained weights on ImageNet, embedding layers are randomly initialized, and the weight of affinity metric is initialized by identity matrix plus random noise.

\textbf{Cross-graph component design}. Our cross-graph affinity component is relatively simple. In fact we also explore a more complex design of cross-graph module, where the matrix $\hat{\mathbf{S}}$ is updated by iterative prediction, rather than predicted from shallower embedding layer as PCA-GM in Alg.~\ref{alg:cross}. In this alternative design, $\hat{\mathbf{S}}^{(0)}$ is initialized as zero matrix, and we iteratively predict $\hat{\mathbf{S}}^{(k)}$ from $\hat{\mathbf{S}}^{(k-1)}$, which is passed to the cross-graph component. Result in Tab.~\ref{tab:pca_iter} reveals that PCA-GM's performance will degrade as $\hat{\mathbf{S}}$ is iteratively predicted, and we further find the training is not stable by this iterative design hence we stick to the simple design in Alg.~\ref{alg:cross}. Details on this alternative design is given in supplementary materials.
\begin{table}[tb!]
\centering
\caption{Accuracy (\%) by number of iterations for a more complex cross-graph affinity component design on Pascal VOC Keypoint, which has negative effect on accuracy (PIA-GM achieves 63.0\%).}
\resizebox{0.48\textwidth}{!}
	{
\begin{tabular}{c|ccccccc|c}
\toprule
 \# of iters & 1 & 2 & 3 & 4 & 5 & 6 & 7 & Alg.~\ref{alg:cross} \\ \midrule 
PCA-GM accuracy & 63.1 & 61.3 & 60.9 & 54.7 & 45.9 & 46.7 & 46.2 & \textbf{63.8}\\
\bottomrule
\end{tabular}
}
%\vspace{1pt}
\vspace{-10pt}
\label{tab:pca_iter}
\end{table}

\textbf{Confusion matrix}. To testify the generalization behavior of our model, we train PCA-GM, PCA-GM-OL, GMN-PL, GMN on eight categories in Pascal VOC Keypoint and report testing result on each category as shown in Fig.~\ref{fig:confusion_matrix}, where result is plotted via confusion matrix ($y$-axis for training and $x$-axis for testing). It shows that embedding adopted in PCA-GM works well, and the permutation loss offers better supervision than the offset one.

\section{Conclusion}
This paper has presented a novel deep learning framework for graph matching, which parameterizes the graph affinity with deep networks and the learning objective involves a permutation loss to account for the arbitrary transformation between two graphs. Extensive experimental results including an ablation study on the presented components and the comparison with peer methods show the state-of-the-art performance of our method.

\newpage
{\small
\bibliographystyle{ieee_fullname}
\bibliography{survey}
}

\end{document}